\documentclass[10pt,twocolumn,letterpaper]{article}

\usepackage{cvpr}
\usepackage{times}
\usepackage{epsfig}
\usepackage{graphicx}
\usepackage{amsmath}
\usepackage{amssymb}
\usepackage{makecell}
\usepackage{enumerate}
\usepackage{algorithm}
\usepackage[noend]{algpseudocode}
\usepackage{booktabs}
\usepackage{subfig}
\usepackage[numbers,sort&compress]{natbib}
\usepackage[usenames, dvipsnames]{color}
\usepackage{color}
\usepackage[normalem]{ulem}
\usepackage[table]{xcolor}
\usepackage{pifont}

\makeatletter
\@namedef{ver@everyshi.sty}{}
\makeatother
\usepackage{pgfplots}
\pgfplotsset{width=70mm,compat=newest}
\newcommand{\red}[1]{\textcolor{red}{#1}}

\newcommand{\cmark}{\color{ForestGreen}\ding{51}}%
\newcommand{\xmark}{\color{red}\ding{55}}

\usepackage[font=small]{caption}

\usepackage[pagebackref=true,breaklinks=true,letterpaper=true,colorlinks,bookmarks=false]{hyperref}
\cvprfinalcopy 

\def\cvprPaperID{2905} 

\begin{document}

\title{Learning without Memorizing}

\author{Prithviraj Dhar*\textsuperscript{1}, Rajat Vikram Singh*\textsuperscript{2}, Kuan-Chuan Peng\textsuperscript{2}, Ziyan Wu\textsuperscript{2}, Rama Chellappa\textsuperscript{1}\\
\textsuperscript{1}University of Maryland, College Park, MD\\
\textsuperscript{2}Siemens Corporate Technology, Princeton, NJ\\
{\tt\small \{prithvi,rama\}@umiacs.umd.edu, \{singh.rajat,kuanchuan.peng,ziyan.wu\}@siemens.com}
}
\maketitle

\begin{abstract}
   Incremental learning (IL) is an important task aimed at increasing the capability of a trained model, in terms of the number of classes recognizable by the model. The key problem in this task is the requirement of storing data (e.g. images) associated with existing classes, while teaching the classifier to learn new classes. However, this is impractical as it increases the memory requirement at every incremental step, which makes it impossible to implement IL algorithms on edge devices with limited memory. Hence, we propose a novel approach, called `Learning without Memorizing (LwM)', to preserve the information about existing (base) classes, without storing any of their data, while making the classifier progressively learn the new classes. In LwM, we present an information preserving penalty: Attention Distillation Loss ($L_{AD}$), and demonstrate that penalizing the changes in classifiers' attention maps helps to retain information of the base classes, as new classes are added. We show that adding $L_{AD}$ to the distillation loss which is an existing information preserving loss consistently outperforms the state-of-the-art performance in the iILSVRC-small and iCIFAR-100 datasets in terms of the overall accuracy of base and incrementally learned classes.
\end{abstract}

\vspace{-1mm}
\section{Introduction}
{\let\thefootnote\relax\footnote{{*These authors have contributed equally to this work, which was partially done during PD's internship at Siemens Corporate Technology. }}}
Most state-of-the-art solutions to visual recognition tasks use models that are specifically trained for these tasks \cite{girshick2015fast,long2015fully}. For the tasks involving categories (such as object classification, segmentation), the complexity of the task (i.e. the number of target classes) limits the ability of these trained models. For example, a trained model aimed for object recognition can only classify object categories on which it has been trained. However, if the number of target classes increases, the model must be updated in such a way that it performs well on the original classes on which it has been trained, also known as base classes, while it incrementally learns new classes as well.

\begin{figure}
{\centering
\includegraphics[width =\linewidth]{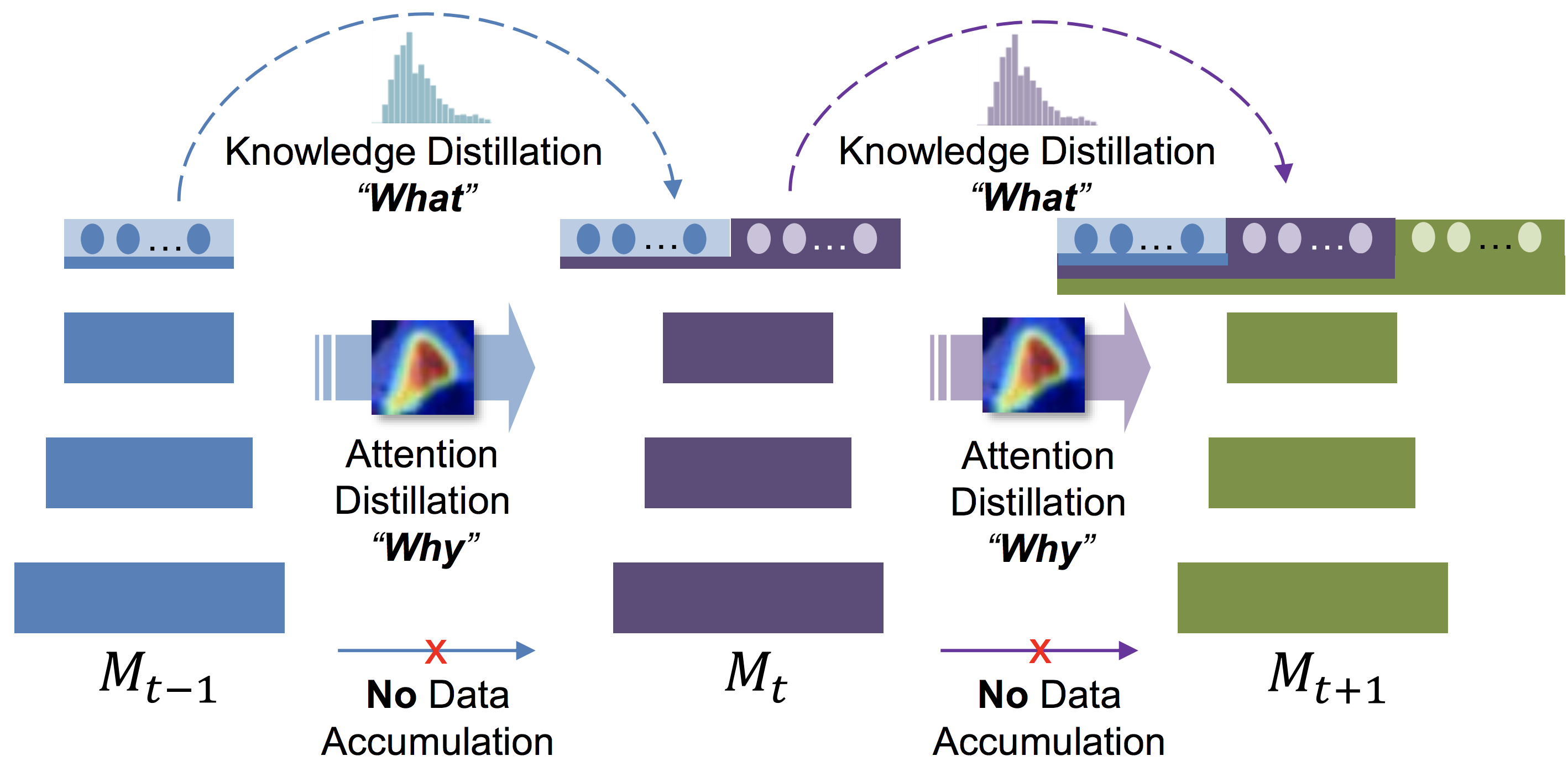}
\caption{Our problem setup does not store data/model pertaining to information about classes learned in previous incremental steps.}
\label{fig:teaser}
\vspace{-1em}
}
\end{figure}
If we retrain the model only on new, previously unseen classes, it would completely forget the base classes, which is known as catastrophic forgetting \cite{kirkpatrick2017overcoming,lee2017overcoming}, a phenomenon which is not typically observed in humane learning. Therefore, most existing solutions \cite{Chaudhry_2018_ECCV,rebuffi2017icarl,wu2018incremental} explore incremental learning (IL) by allowing the model to retain a fraction of the training data of base classes, while incrementally learning new classes. Yu et al. \cite{wu2018incremental} proposed retaining trained models encoding base class information, to transfer their knowledge to the model learning new classes. However, this process is not scalable. This is because storing base class data or models encoding base class information is a memory expensive task, and hence is cumbersome when used in a lifelong learning setting. Also, in an industrial setting, when a trained object classification model is delivered to the end user, the training data is kept private for proprietary reasons. Therefore, the end user will be unable to update the trained model to incorporate new target classes in the absence of base class data.

Moreover, storing base class data for incrementally learning new classes is not biologically inspired. For example, when a toddler learns to recognize new shapes/objects, it is observed that it does not completely forget the shapes or objects it already knows. It also does not always need to revisit the old information when learning new entities. Inspired by this, we aim to explore incremental learning in object classification by adding a stream of new classes without storing data belonging to classes that the classifier has already seen. While IL solutions that do not require base class data, such as \cite{kirkpatrick2017overcoming,aljundi2017memory} have been proposed, these methods mostly aim at incrementally learning new tasks, which means that at test time the model cannot confuse the incrementally learned tasks with tasks it has already learned, making the problem setup much easier.

We explore the problem of incrementally learning object classes, without storing any data or model associated with the base classes (Figure \ref{fig:teaser}) in the previous steps, while allowing the model to confuse new classes with old ones. In our problem setup, an ideal incremental learner should have the following properties:
\begin{enumerate}[i]
  \item It should help a trained model to learn new classes obtained from a stream of data, while preserving the model's knowledge of base class information.
  \item At testing time, it should enable the model to consider all the classes it has learned when the model makes a prediction.
  \item The size of the memory footprint should not grow at all, irrespective of the number of classes seen thus far.
\end{enumerate}
An existing work targeting the same problem is LwF-MC, which is one of the baselines in \cite{rebuffi2017icarl}. In the following sections, we use the following terminology (introduced in \cite{zagoruyko2016paying}) at incremental step $t$ :\\
\textbf{Teacher model, $M_{t-1}$}, i.e. the model trained with only base classes.\\ \textbf{Student model, $M_t$}, i.e. the model which incrementally learns new classes, while emulating the teacher model for maintaining performance on base classes.\\
\textbf{Information Preserving Penalty (IPP)}, i.e. the loss to penalize the divergence between $M_{t-1}$ and $M_t$. Ideally, this helps $M_t$ to be as proficient in classifying base classes as $M_{t-1}$.

Initialized using $M_{t-1}$, $M_t$ is then trained to learn new classes using a classification loss, $L_C$. However, an IPP is also applied to $M_t$ so as to minimize the divergence between the representations of $M_{t-1}$ and $M_t$. While $L_C$ helps $M_t$ to learn new classes, IPP prevents $M_t$ from diverging too much from $M_{t-1}$. Since $M_t$ is already initialized as $M_{t-1}$, the initial value of IPP is expected to be close to zero. However, as $M_t$ keeps learning new classes with $L_C$, it starts diverging from $M_{t-1}$, which leads the IPP to increase. The purpose of the IPP is to prevent the divergence of $M_t$ from $M_{t-1}$. Once $M_t$ is trained for a fixed number of epochs, it is used as a teacher in the next incremental step, using which a new student model is initialized.

\begin{figure}
\centering
{\includegraphics[width=\linewidth]{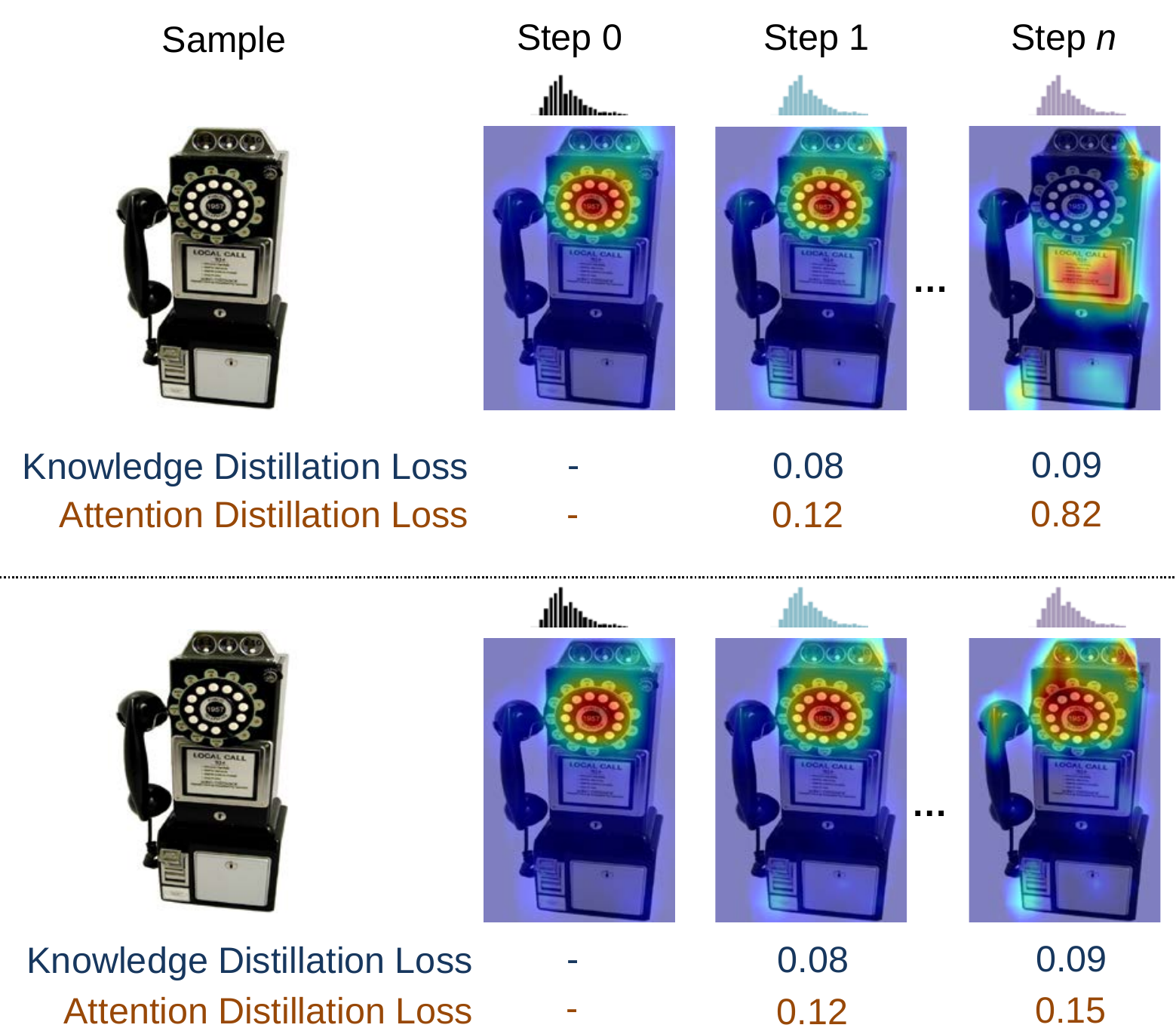}}
\vspace{-1.5em}
\caption{(Top) Example of a case where attention regions degrade in later incremental steps.(Bottom) Example of a case where attention regions do not vary across incremental steps. Distillation loss is seen to be unaffected by degrading attention regions, whereas Attention Distillation Loss is sensitive to the attention regions}
\label{fig:iou_ad}
\vspace{-.5em}
\end{figure}

In LwF-MC \cite{rebuffi2017icarl}, the IPP is the knowledge distillation loss. The knowledge distillation loss $L_D$, in this context, was first introduced in \cite{li2017learning}. It captures the divergence between the prediction vectors of $M_{t-1}$ and $M_t$. In an incremental setup, when an image belonging to a new class ($I_n$) is fed to $M_{t-1}$, the base classes which have some resemblance in $I_n$ are captured. $L_D$ enforces $M_t$ to capture the same base classes. Thus, $L_D$ essentially makes $M_t$ learn `what' are the possible base classes in $I_n$, as shown in Figure \ref{fig:teaser}. Pixels that have significant influence on the models' prediction constitute the attention region of the network. However, $L_D$ does not explicitly take into account the degree of each pixel influencing the models’ predictions. For example, in Figure \ref{fig:iou_ad}, in the first row, it is seen that at step $n$, even though the network focuses on an incorrect region while predicting `dial\_telephone', the numerical value of $L_D$ (0.09) is same as that when the network focuses on the correct region in step $n$, in the bottom row.

We hypothesize that attention regions encode the models' representation more precisely. Hence, constraining the attention regions of $M_t$ and $M_{t-1}$ using an Attention Distillation Loss ($L_D$, explained in Sec. \ref{sec:adl}), to minimize the divergence of the representations of $M_t$ from that of $M_{t-1}$ is more meaningful. This is because, instead of finding which base classes are resembled in the new data, attention maps explain `why' hints of a base class are present (as shown in Figure \ref{fig:teaser}). Using these hints, $L_D$, in an attempt to make the attention maps of $M_{t-1}$ and $M_t$ equivalent, helps to encode some visual knowledge of base class in $M_t$. We show the utility of $L_{AD}$ in Figure \ref{fig:iou_ad}, where although the model correctly predicts the image as 'dial\_telephone', the value of $L_D$ in step $n$ increases if the attention regions diverge too much from the region in Step 0.

We propose an approach where an Attention Distillation Loss ($L_{AD}$) is applied to $M_t$ to prevent its divergence from $M_{t-1}$, at incremental step $t$. Precisely, we propose to constrain the $L_1$ distance between the attention maps generated by $M_{t-1}$ and $M_t$ in order to preserve the knowledge of base classes. The reasoning behind this strategy is described in Sec \ref{sec:adl}. This is applied in addition to the distillation loss $L_D$ and a classification loss for the student model to incrementally learn new classes.

The main contribution of this work is to provide an attention-based approach, termed as `Learning without Memorizing (LwM)',  that helps a model to incrementally learn new classes by restricting the divergence between student and teacher model. LwM does not require any data of the base classes when learning new classes. Different from contemporary approaches that explore the same problem, LwM takes into account the gradient flow information of teacher and student models by generating attention maps using these models. It then constrains this information to be equivalent for teacher and student models, thus preventing the student model to diverge too much from the teacher model. Finally, we show that LwM consistently outperforms the state-of-the-art performance in the iILSVRC-small \cite{rebuffi2017icarl} and iCIFAR-100 \cite{rebuffi2017icarl} datasets.

\section{Related work}

In object classification, Incremental learning (IL) is the process of increasing the breadth of an object classifier, by training it to recognize new classes, while retaining its knowledge of the classes on which it has been trained originally. In the past couple of years, there has been considerable research efforts in this field \cite{kirkpatrick2017overcoming,li2017learning}. Moreover, there exist several subsets of this research problem which impose different constraints in terms of data storage and evaluation. We can divide existing methods based on their constraints:

\textbf{Task incremental (TI) methods:} In this problem, a model trained to perform object classification on a specific dataset is incrementally trained to classify objects in a new dataset. A key characteristic of these experiments is that during evaluation, the final model is tested on different datasets (base and incrementally learned) separately. This is known as multi-headed evaluation \cite{Chaudhry_2018_ECCV}. In such an evaluation, the classes belonging to two different tasks have no chance to confuse with one another. One of the earlier works in this category is LwF \cite{li2017learning}, where a distillation loss is used to preserve information of the base classes. Also, the data from base classes is used during training, while the classifier learns new classes.  A prominent work in this area is EWC \cite{kirkpatrick2017overcoming}, where at each incremental task the weights of the student model are set to those of their corresponding teacher model, according to their importance of network weights. Aljundi et al. present MAS \cite{aljundi2017memory}, a technique to train the agents to learn what information should not be forgotten. All experiments in this category use multi-headed evaluation, which is different from the problem setting of this paper where we use single-headed evaluation, defined explicitly in \cite{Chaudhry_2018_ECCV}. Single-headed evaluation is another evaluation method wherein the model is evaluated on both base and incrementally learned classes jointly. Multi-headed evaluation is easier than single-headed evaluation, as explained in \cite{Chaudhry_2018_ECCV}.

\textbf{Class incremental (CI) methods:} In this problem, a model trained to perform object classification on specific classes of a dataset is incrementally trained to classify new unseen classes in the same dataset. Most of the existing work exploring this problem use single-headed evaluation. This makes the CI problem more difficult than the TI problem because the model can confuse the new class with a base class in the CI problem. iCaRL \cite{rebuffi2017icarl} belongs to this category. In iCaRL \cite{rebuffi2017icarl}, Rebuffi et al. propose a technique to jointly learn feature representation and classifiers. They also introduce a strategy to select exemplars which is used in combination with the distillation loss to prevent catastrophic forgetting. In addition, a new baseline: LwF-MC is introduced in \cite{rebuffi2017icarl}, which is a class incremental version of LwF \cite{li2017learning}. LwF-MC uses the distillation loss to preserve the knowledge of base classes along with a classification loss, without storing the data of base classes and is evaluated using single-headed evaluation. Another work aiming to solve the CI problem is \cite{Chaudhry_2018_ECCV}, which evaluates using both single-headed and multi-headed evaluations and highlights their difference. Chaudhry et al. \cite{Chaudhry_2018_ECCV} introduce metrics to quantify forgetting and intransigence, and also propose the Riemannian walk to incrementally learn classes.

A key factor of most incremental learning frameworks is whether or not they allow storing the data of base classes (i.e. classes on which the classifier is originally trained). We can also divide existing methods based on this factor:

\textbf{Methods which use base class data:} Several experiments have been proposed to use a small percentage of the data of base classes while training the classifier to learn new classes. iCaRL \cite{rebuffi2017icarl} uses the exemplars of base classes, while incrementally learning new classes. Similarly, Chaudhry et al. \cite{Chaudhry_2018_ECCV} also use a fraction of the data of base classes. Chaudhry et al. \cite{Chaudhry_2018_ECCV} also show that this is especially useful for alleviating intransigence, which is a problem faced in single-headed evaluation. However, storing data for base classes increases memory requirement at each incremental step, which is not feasible when the memory budget is limited.

\textbf{Methods which do not use base class data: } Several TI methods described earlier (such as \cite{aljundi2017memory, kirkpatrick2017overcoming} ) do not use the information about base classes while training the classifier to learn new classes incrementally. To the best of our knowledge, LwF-MC \cite{rebuffi2017icarl} is the only CI method which needs no base class data but uses single-headed evaluation.

\begin{table}
\centering
\small{
\begin{tabular}{ccc}
\toprule
 Constraints & Use base class data & No base class data  \\
\midrule
CI methods &iCaRL \cite{rebuffi2017icarl}, \cite{Chaudhry_2018_ECCV}, \cite{wu2018incremental} &\cellcolor{blue!30} LwF-MC \cite{rebuffi2017icarl}, \textbf{LwM}\\
\midrule
TI methods &LwF \cite{li2017learning} &\thead{IMM \cite{lee2017overcoming}, EWC \cite{kirkpatrick2017overcoming}, \\MAS \cite{aljundi2017memory}, \cite{aljundi2018selfless}, \cite{jung2016less}}\\
\bottomrule
\end{tabular}
\vspace{-.5em}
\caption{Categorization of recent related works in incremental learning. We focus on the class incremental (CI) problems where base class data is unavailable when learning new classes.}
\label{table:rw}
\vspace{-.5em}
}
\end{table}
Table \ref{table:rw} presents a taxonomy of previous works in this field. We propose a technique to solve the CI problem, without using any base class data. We can infer from the discussion above that LwF-MC \cite{rebuffi2017icarl} is the only existing work which uses single-headed evaluation, and hence use it as our baseline. We intend to use attention maps in an incremental setup, instead of only knowledge distillation, to transfer more comprehensive knowledge of base classes from teacher to student model. Although in \cite{zagoruyko2016paying}, enforcing equivalence of attention maps of teacher and student models has been explored previously for transferring knowledge from teacher to student models, the same approach cannot be applied to an incremental learning setting. In our incremental problem setup, due to the absence of base class data, we intend to utilize the attention region in the new data which resembles one of the base classes. But these regions are not prominent since the data does not belong to any of the base classes, thus making class-specific attention maps a necessity. Class-specificity is required to mine out base class regions in a more targeted fashion, which is why generic attention maps such as activation-based attention maps in \cite{zagoruyko2016paying} are not applicable as they can not provide a class-specific explanation about relevant patterns corresponding to the target class. We define class-specific interpretation as how a network understands the spatial locations of specific kinds of object. Such locations are determined by computing Grad-CAM \cite{selvaraju2017grad} attention maps. Also, in LwM, by using class-specific attention map, we can enforce the consistency on class-specific interpretation between teacher and student models. Moreover, our problem setup is different from knowledge distillation because at incremental step $t$, we freeze $M_{t-1}$ while training $M_t$, and do not allow $M_t$ to access data from the base classes, and therefore  $M_{t-1}$ and $M_t$ are trained using a completely different set of classes. This makes the problem more challenging as the output of $M_t$ on feeding data from unseen classes is the only source of base class data. This is further explained in Sec. \ref{sec:adl}.

We intend to explore the CI problem by proposing to constrain the attention maps of the teacher and student models to be equivalent (in addition to their prediction vectors), to improve the information preserving capability of LwF-MC \cite{rebuffi2017icarl}. In LwF-MC and our proposed method LwM, storing teacher models trained in previous incremental steps is not allowed since it would not be feasible to accumulate models from all the previous steps when the memory budget is limited.

\section{Background}
 Before we discuss LwM, it is important to introduce distillation loss $L_D$, which is our baseline IPP, as well as how we generate attention maps.


\subsection{Distillation loss ($L_D$)}
 $L_D$ was first introduced in \cite{li2017learning} for incremental learning. It is defined as follows:
\begin{equation}
\label{eq:ld}
    L_D(\mathbf{y},\mathbf{\hat{y}})=-\sum_{i=1}^{N} y^{'}_i.\log(\hat{y^{'}_i}),
\end{equation}
where $\mathbf{y}$ and $\mathbf{\hat{y}}$ are prediction vectors (composed of probability scores) of $M_{t-1}$ and $M_t$ for base classes at incremental step $t$, each of length $N$ (assuming that $M_{t-1}$ is trained on $N$ base classes). Also, $y^{'}_i=\sigma(y_i)$ and $\hat{y^{'}_i}=\sigma(\hat{y_i})$ (where $\sigma(\cdot)$ is sigmoid activation). This definition of $L_D$ is consistent with that defined in LwF-MC \cite{rebuffi2017icarl}. Essentially, $L_D$ enforces the base class prediction of $M_t$ and $M_{t-1}$ to be equivalent, when an image belonging to one of the incrementally added classes is fed to each of them. Moreover, we believe that there exist common visual semantics or patterns in both base and new class data. Therefore, it makes sense to encourage the feature responses of $M_t$ and $M_{t-1}$ to be equivalent, when new class data is given as input. This helps to retain the old class knowledge (in terms of the common visual semantics).

\subsection{Generating attention maps}
We describe the technique employed to generate attention maps. In our experiments we use the Grad-CAM \cite{selvaraju2017grad} for this task. In \cite{selvaraju2018choose}, Grad-CAM maps have been shown to encode information to learn new classes, although not in an incremental setup. For using the Grad-CAM, the image is first forwarded to the model, obtaining a raw score for every class. Following this, the gradient of score $y^c$ for a desired class $c$ is computed with respect to each convolutional feature map $A_k$. For each $A_k$, global average pooling is performed to obtain the neuron importance $\alpha_k$ of $A_k$. All the $A_k$ weighted by $\alpha_k$ are passed through a ReLU activation function to obtain a final attention map for class $c$.

More precisely, let $\alpha_k = \frac{\partial y^c}{\partial A_k} $. Let $\mathbf{\alpha}= [\alpha_1, \alpha_2, \dots, \alpha_K ]$ and $\mathbf{A}=[A_1, A_2, \dots, A_K ] $, where $K$ is the number of convolutional feature maps in the layer using which attention map is to be generated. The attention map $Q$ can be defined as
\vspace{-0.5em}
\begin{equation}
\label{eq:att}
    Q = ReLU(\mathbf{\alpha}^T\mathbf{A})
\vspace{-0.5em}
\end{equation}

\begin{figure}
\centering
{\includegraphics[width = 1\linewidth]{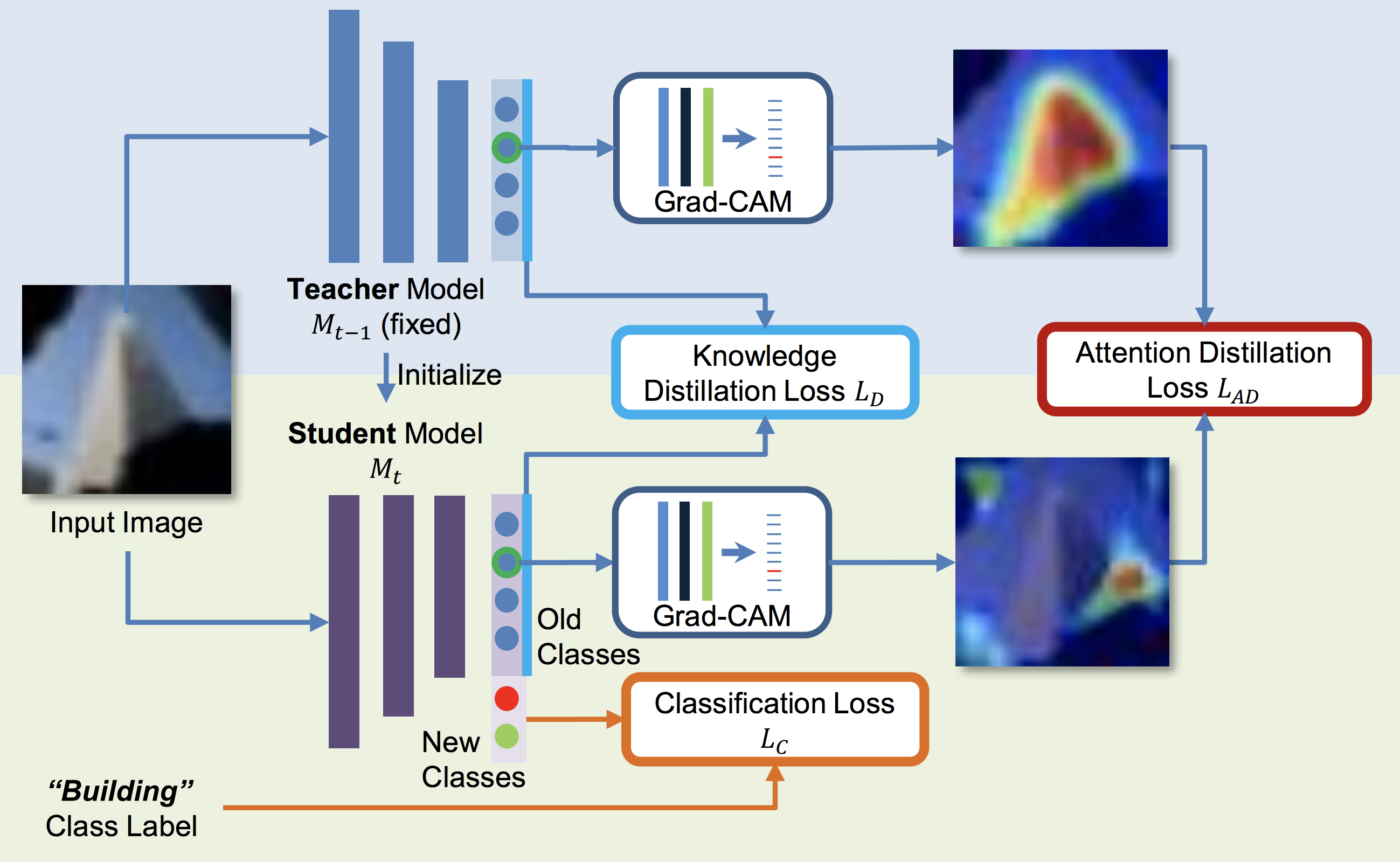}}
\caption{At incremental step $t$, LwM accepts images belonging to one of the new classes. Three losses ($L_C, L_D \text{ and $L_{AD}$}$) are applied to $M_t$ while $M_{t-1}$ remains frozen. The new classes are depicted in the lower part of the classifier of $M_t$.}
\label{fig:visual}
\end{figure}
\section{Proposed approach}
We introduce an information preserving penalty ($L_{AD}$) based on attention maps. We combine $L_{AD}$ with distillation loss $L_D$ and a classification loss $L_C$ to construct LwM, an approach which encourages attention maps of teacher and student to be similar. Our LwM framework is shown in Figure \ref{fig:visual}. The loss function of LwM is defined below:
\begin{equation}
\label{eq:lwm}
    L_{LwM}= L_C + \beta L_D +\gamma L_{AD}
\end{equation}
Here $\beta,\gamma$ are the weights used for $L_D, \text{$L_{AD}$}$ respectively. In comparison to LwM, LwF-MC \cite{rebuffi2017icarl} only uses a classification loss combined with distillation loss and is our baseline.

\subsection{Attention distillation loss ($L_{AD}$)}
\label{sec:adl}
At incremental step $t$, we define student model $M_t$, initialized using a teacher model $M_{t-1}$. We assume $M_t$ is proficient in classifying $N$ base classes. $M_t$ is required to recognize $N+k$ classes, where $k$ is the number of previously unseen classes added incrementally. Hence, the sizes of the prediction vectors of $M_{t-1}$ and $M_t$ are $N$ and $N+k$ respectively.  For any given input image $i$, we denote the vectorized attention maps generated by $M_{t-1}$ and $M_{t}$, for class $c$ as $Q^{i,c}_{t-1}$ and $Q^{i,c}_t$, respectively. We generate these maps using Grad-CAM \cite{selvaraju2017grad}, as explained above.
\begin{equation}
    Q^{i,c}_{t-1} = vector(\text{Grad-CAM}(i,M_{t-1},c))
\end{equation}
\begin{equation}
    Q^{i,c}_t =  vector(\text{Grad-CAM}(i,M_t,c))
\end{equation}
 We assume that the lengths of each vectorized attention map is $l$. In \cite{zagoruyko2016paying}, it has been mentioned that normalizing the attention map by dividing it by the $L_2$ norm of the map is an important step for student training. Hence we perform this step while computing $L_{AD}$. During training of $M_t$, an image belonging to one of the new classes to be learned (denoted as $I_n$), is given as input to both $M_{t-1}$ and $M_t$. Let $b$ be the top base class predicted by $M_t$ (i.e. base class having the highest score) for $I_n$.  For this input, $L_{AD}$ is defined as the sum of element wise $L_1$ difference of the normalized, vectorized attention map:
\begin{equation}
\label{eq:adl}
    \text{$L_{AD}$} = \sum_{j=1}^l\|\frac{Q^{I_n, b}_{t-1,j}}{\|Q^{I_n, b}_{t-1}\|_2} - \frac{Q^{I_n, b}_{t,j}}{\|Q^{I_n, b}_{t}\|_2}\|_1
\end{equation}
From the explanation above, we know that for training $M_{t}$, $M_{t-1}$ is fed with the data from the classes that it has not seen before ($I_n$). Essentially, the attention regions generated by $M_{t-1}$ for $I_n$, represent the regions in the image which resemble the base classes. If $M_t$ and $M_{t-1}$ have equivalent knowledge of base classes, they should have a similar response to these regions, and therefore $Q^{I_n, b}_{t}$ should be similar to $Q^{I_n, b}_{t-1}$. This implies that the attention outputs of $M_{t-1}$ are the only traces of base data, which guides $M_t$'s knowledge of base classes. We use the $L_1$ distance between $Q^{I_n, b}_{t-1}$ and $Q^{I_n, b}_t$ as a penalty to enforce their similarity. We experimented with both $L_1$ and $L_2$ distance in this context. However, as we obtained better results with $L_1$ distance on  held-out data, we chose $L_1$ over $L_2$ distance.

According to Eq. \ref{eq:att}, attention maps encode gradient of the score of class $b$, $y^b$ with respect to convolutional feature maps $\mathbf{A}$. This information is not explicitly captured by the distribution of class scores (used by $L_D$). By encouraging $Q^{I_n,b}_{t-1}$ and $Q^{I_n,b}_t$ to be equivalent, we are restricting the divergence between $\Bigg[\frac{\partial y^{b}}{\partial \mathbf{A}}\Bigg]_{t-1}$ and $\Bigg[\frac{\partial y^{b}}{\partial \mathbf{A}}\Bigg]_t$. This ensures the consistency on class-specific interpretation between teacher and student. We know that every feature map in $\mathbf{A}$ encodes a visual feature. While there can be several factors that can cause changes to $y^b$, $L_{AD}$ forces the changes with respect to a specific visual feature encapsulated in $\mathbf{A}$ to be equivalent for $M_t$ and $M_{t-1}$. Hence, we hypothesize that combining $L_D$, which captures the score distribution of the model for base classes ($\mathbf{y},\mathbf{\hat{y}}$), with a loss that captures the gradient flow information of the model, would result in a more wholesome information preserving penalty. Moreover, the attention maps are a 2D manifestation of the prediction vectors ($\mathbf{y},\mathbf{\hat{y}}$), which means that they capture more spatial information than these vectors, and hence it is more advantageous to use attention maps than using only prediction vectors.

\begin{figure*}
\centering
{\includegraphics[width = \linewidth]{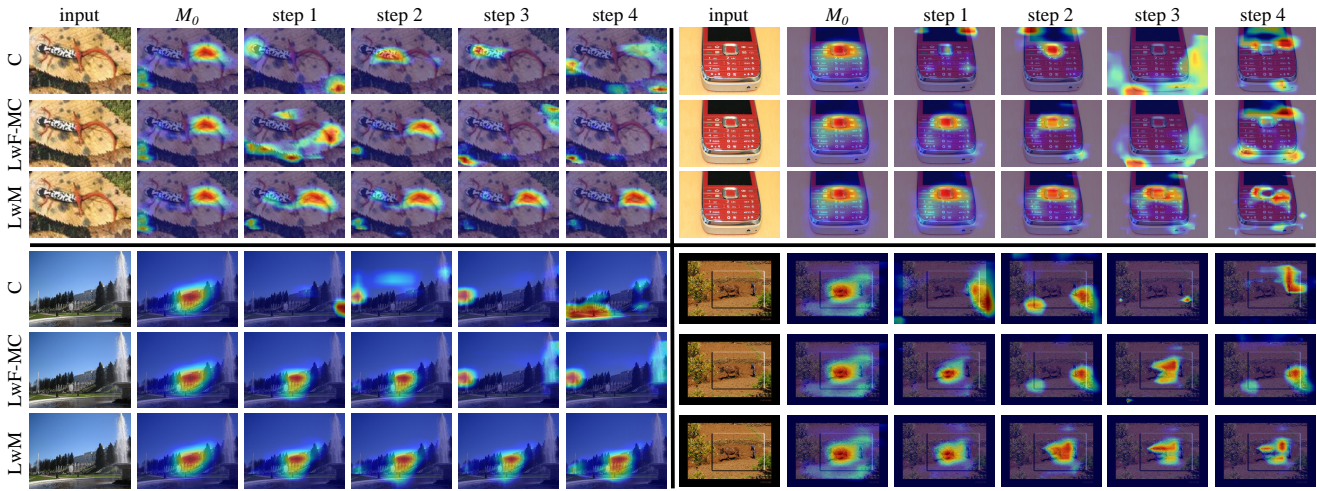}}
\vspace{-1.8em}
\caption{The example attention maps generated by the following experiment IDs (Table~\ref{table:para}): C, LwF-MC, and LwM. All the input images belong to the initial base classes. The column $M_0$ represents the corresponding base-class attention maps generated by the initial teacher model, and the columns step 1$\sim$4 represent the corresponding base-class attention maps generated in four different incremental steps in temporal order. These examples show that the attention maps generated by LwM are closer to those in the column $M_0$ over time compared with C and LwF-MC, which demonstrates the efficacy of $L_{AD}$ in LwM.}
\label{fig:att_over_time}
\vspace{-.5em}
\end{figure*}

\section{Experiments}
We first explain our baseline, which is LwF-MC \cite{rebuffi2017icarl}. Following that, we provide information about the datasets used in our experiments. After that, we describe the iterative protocol to perform classification at every incremental step. We also provide implementation details including architectural information.
\subsection{Baseline}
As our baseline is LwF-MC \cite{rebuffi2017icarl}, we firstly implement its objective function, which is a sum of a classification loss and distillation loss ($L_C + L_D$). In all our experiments, we use a cross entropy loss for $L_C$ to be consistent with \cite{rebuffi2017icarl}. However, it should be highlighted that the official implementation of $L_D$ in LwF-MC by \cite{rebuffi2017icarl} is different from the definition of $L_D$ in \cite{li2017learning}. As LwF-MC (but not LwF) is our baseline, we use iCaRL's implementation of LwF-MC in our work. LwF cannot handle CI problems where no base class training data is available (according to Table \ref{table:rw}), which is the reason why we choose LwF-MC as the baseline and iCaRL's implementation.

\begin{table}
\centering
\small{
\begin{tabular}{@{\hspace{0mm}}c@{\hspace{-1mm}}c@{\hspace{2mm}}c@{\hspace{2mm}}c@{\hspace{2mm}}c}
\toprule
Dataset &iILSVRC &iCIFAR &CUB200 & Caltech\\
 & -small & -100 & -2011 & -101\\
\midrule
\# classes &100 &100 &100 & 100\\
\midrule
\# training images &500 &500 &80\% of data& 80\% of data\\
\# testing images &100 &100 &20\% of data& 20\% of data\\
\# classes/batch &10 &10, 20, 50 &10 & 10\\
eval. metric &top-5 &top-1 &top-1 &top-1 \\
\bottomrule
\end{tabular}
\vspace{-.5em}
\caption{The statistics of the datasets used in our experiments, in accordance with \cite{rebuffi2017icarl}. Additionally, we also perform experiments on the CUB-200-2011 \cite{WahCUB_200_2011} dataset.}
\label{table:dataset_all}
\vspace{-.5em}
}
\end{table}

\subsection{Datasets}
We use two datasets used in LwF-MC \cite{rebuffi2017icarl} for our experiments. Additionally, we also perform experiments on Caltech-101 \cite{FeiFei2006caltech101} as well as CUBS-200-2011 \cite{WahCUB_200_2011} datasets.  The details for the datasets are provided in Table \ref{table:dataset_all}. These datasets are constructed by randomly selecting a batch of classes at every incremental step. In both datasets, the classes belonging to different batches are disjoint. For a fair comparison, the data preparation for all the datsets and evaluation strategy are the same as that for LwF-MC \cite{rebuffi2017icarl}.

\subsection{Experimental protocol}
\label{sec:prot}
We now describe the protocol using which we iteratively train $M_t$, so that it preserves the knowledge of the base classes while incrementally learning new classes.

\textbf{Initialization: } Before the first incremental step ($t=1$), we train a teacher model $M_0$ on 10 base classes, using a classification loss for 10 epochs. The classification loss is a cross entropy loss $L_C$. Following this, for $t=1$ to $t=k$ we initialize student $M_t$ using $M_{t-1}$ as its teacher, and feed data from a new batch of images that is to be incrementally learned, to both of these models. Here $k$ is the number of incremental steps.

\textbf{Applying IPP and classification loss to student model: }Given the data from new classes as inputs, we generate the output of $M_t$ and $M_{t-1}$ with respect to base class having the highest score. These outputs can either be class-specific attention maps (required for computing $L_{AD}$) or class-specific scores (required for computing $L_D$). Using these outputs we compute an IPP which can either be $L_{AD}$ or $L_D$. In addition, we apply a classification loss to $M_t$ based on its outputs with respect to the new classes which are to be learned incrementally. We jointly apply classification loss and IPP to $M_t$ and train it for 10 epochs. Once $M_t$ is trained, we use it as a teacher model in the next incremental step, and follow the aforementioned steps iteratively, until all the $k$ incremental steps are completed.

\begin{table}
\centering
\small{
\begin{tabular}{cccc}
\toprule
Experiment ID\textbackslash loss &$L_C$ &$L_D$ &$L_{AD}$\\
\midrule
Finetuning &\cmark&\xmark &\xmark\\
LwF-MC \cite{rebuffi2017icarl} &\cmark &\cmark &\xmark\\
LwM &\cmark &\cmark &\cmark\\
\bottomrule
\end{tabular}
\vspace{-.5em}
\caption{Experiment configurations used in this work, identified by their respective experiment IDs.}
\label{table:para}
\vspace{-.5em}
}
\end{table}
\subsection{Implementation details}

We use the ResNet-18 \cite{he2016deep} architecture for training student and teacher models on the iILSVRC-small, Caltech-101 and CUBS-200-2011 datasets, and the ResNet-34 \cite{he2016deep} for training models on the iCIFAR-100 dataset. This is consistent with the networks and datasets used in \cite{rebuffi2017icarl}. We used a learning rate of 0.01. The feature maps of the final convolutional layer are used to generate attention maps using Grad-CAM, as these maps are highly interpretable. \cite{selvaraju2017grad}. The combinations of classification loss and IPP, along with their experiment IDs are provided in Table \ref{table:para}. The experiment configurations will be referred to as their respective experiment IDs from now on.
\section{Results}

\begin{figure*}
\centering
{\includegraphics[width = \linewidth]{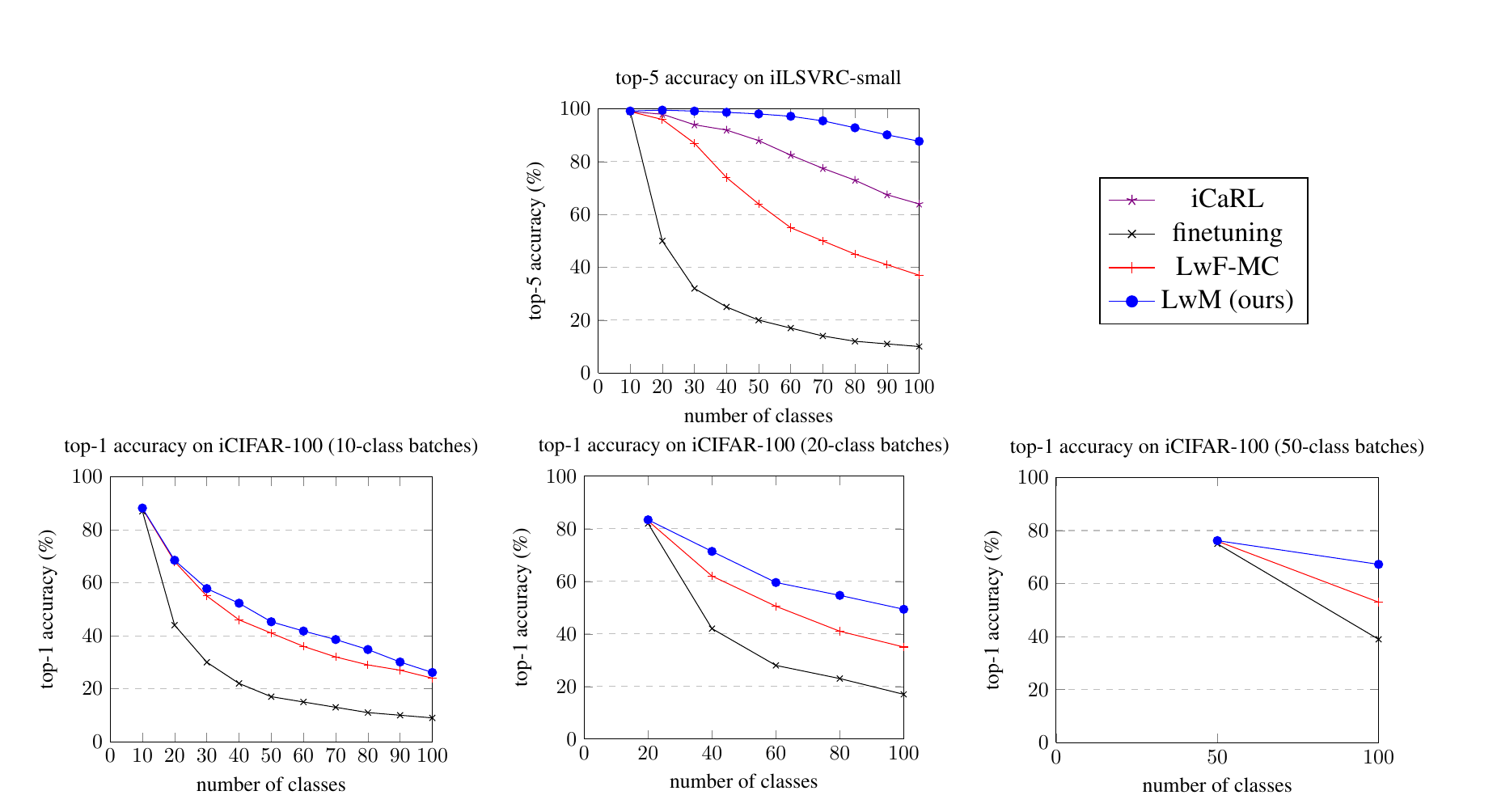}}
\vspace{-1.9em}
\caption{Performance comparison between our method, LwM, and the baselines. LwM outperforms LwF-MC \cite{rebuffi2017icarl} and ``using only classification loss with finetuning" on the iILSVRC-small and iCIFAR-100 datasets \cite{rebuffi2017icarl}. LwM even outperforms iCaRL \cite{rebuffi2017icarl} on the iILSVRC-small dataset given that iCaRL has the unfair advantage of accessing the base-class data.
}
\label{fig:results}
\end{figure*}

Before discussing the quantitative results and advantages of our proposed penalties, we show some qualitative results to demonstrate the advantage of using $L_{AD}$. We show that we can retain attention regions of base classes for a longer time when more classes are incrementally added to the classifier by using LwM as compared to LwF-MC \cite{rebuffi2017icarl}. Before the first incremental step $t=1$, we have $M_0$ trained on 10 base classes. Now, following the protocol in Sec. \ref{sec:prot}, we incrementally add 10 classes at each incremental step. At every incremental step $t$, we train $M_t$ with 3 configurations: C, LwF-MC \cite{rebuffi2017icarl}, and LwM. We use $M_t$ to generate the attention maps for the data from base classes (using which $M_0$ was trained), which it has not seen, and show the results in Figure \ref{fig:att_over_time}. Additionally, we also generate corresponding attention maps using $M_0$ (i.e. the first teacher model), which can be considered `ideal' (as target maps) as $M_0$ was given full access to base class data. For the $M_t$s trained with C, it is seen that attention regions for base classes are quickly forgotten after every incremental step. This can be attributed to catastrophic forgetting \cite{kirkpatrick2017overcoming,lee2017overcoming}. $M_t$ trained with LwF-MC \cite{rebuffi2017icarl} have slightly better attention preserving ability but as the number of incremental steps increases, the attention regions diverge from the `ideal' attention regions. Interestingly, the attention maps generated by $M_t$ trained with LwM configuration retain the attention regions for base classes for all incremental steps shown in Figure \ref{fig:att_over_time}, and are most similar to the target attention maps. These examples support that LwM delays forgetting of base class knowledge.

We now present the quantitative results of the following configurations: C , LwF-MC \cite{rebuffi2017icarl} and LwM. To show the efficacy of LwM across, we evaluate these configurations on multiple datasets. The results on the iILSVRC-small and iCIFAR-100 datasets are presented in Figure \ref{fig:results}. For the iILSVRC-small dataset, the performance of LwM is better than that of the baseline LwF-MC \cite{rebuffi2017icarl}. LwM outperforms the baseline by a margin of more than 30\% when the number of classes is 40 or more. Especially for 100 classes, LwM achieves an improvement of more than 50\% over the baseline LwF-MC \cite{rebuffi2017icarl}. In addition, LwM outperforms iCaRL \cite{rebuffi2017icarl}, at every incremental step, even though iCaRL has the unfair advantage of storing the exemplars of base classes while training the student model for the iILSVRC-small dataset.

\begin{table}
\centering
\small{
\begin{tabular}{c|cc|cc}
\toprule
\# Classes & FT & LwM (ours)& FT & LwM (ours)\\
\midrule
Dataset & \multicolumn{2}{c|}{Caltech-101} & \multicolumn{2}{c}{CUBS-200-2011} \\
\midrule
10 (base)  &97.78	&97.78	&99.17	&99.17 \\
20  &59.55	&\textbf{75.34}	&57.92	&\textbf{78.75} \\
30  &52.65	&\textbf{71.78}	&41.11	&\textbf{70.83} \\
40 	&44.51	&\textbf{67.49}	&35.42	&\textbf{58.54} \\
50	&35.52	&\textbf{59.79}	&32.33	&\textbf{53.67} \\
60	&31.18	&\textbf{56.62}	&29.03	&\textbf{47.92} \\
70	&32.99	&\textbf{54.62}	&22.14	&\textbf{43.79} \\
80	&27.45	&\textbf{48.71}	&22.27	&\textbf{43.83} \\
90	&28.55	&\textbf{46.21}	&20.52	&\textbf{39.85} \\
100	&28.26	&\textbf{48.42}	&17.4	&\textbf{34.52} \\
\bottomrule
\end{tabular}
\vspace{-.5em}
\caption{Results obtained on Caltech-101 \cite{FeiFei2006caltech101} and CUBS-200-2011 \cite{WahCUB_200_2011}. Here FT refers to finetuning. The first step refers to the training of first teacher model using 10 classes. }
\label{table:calcub}
\vspace{-.5em}
}
\end{table}

\begin{table}
\renewcommand\thetable{5}
\centering
\small{
\begin{tabular}{cccc}
\toprule
\# Classes / Config & $L_C+L_{AD}$  & LwM (ours)   \\
\midrule
20 & 84.95&  \bf{99.55}\\
30 & 55.82& \bf{99.18}\\
40 & 43.46&  \bf{98.72}\\
50 & 36.36&  \bf{98.10}\\
60 & 26.78& \bf{97.22}\\
\bottomrule
\end{tabular}
\vspace{-0.5em}
\caption{Top-5 accuracy comparison of $L_C$ + $L_{AD}$ and LwM. The LwM accuracies are in accordance to that of Figure \ref{fig:results}. Not designed to be used alone, $L_{AD}$ is used to ensure the consistency on class-specific interpretation between teacher and student, by enforcing the gradients of class-specific score w.r.t. feature maps to be equivalent.}
\label{table:5}
\vspace{-0.5em}
}
\end{table}
To be consistent with the LwF-MC experiments in \cite{rebuffi2017icarl}, we perform experiments by constructing the iCIFAR-100 datasets by using batches of 10, 20, and 50 classes at each incremental step. The results are provided in Figure \ref{fig:results}. It can be seen that LwM outperforms LwF-MC for all three sizes of incremental batches in iCIFAR-100 dataset. Hence, we conclude that LwM consistently outperforms LwF-MC \cite{rebuffi2017icarl} in iILSVRC-small and iCIFAR-100 datasets. Additionally, we also perform these experiments using the Caltech-101 and CUBS-200-2011 dataset \cite{FeiFei2006caltech101} by adding a batch of 10 classes at every incremental step and compare it with finetuning. The results for these two datasets are shown in Table \ref{table:calcub}.In Table \ref{table:5}, we also provide the results obtained using only a combination of $L_C$ and $L_{AD}$, on a few incremental steps in iILSVRC-small dataset.

The advantage of incrementally adding every loss on top of $L_C$ is demonstrated in Figure \ref{fig:results}, where we show that the performance with only C is poor due to the catastrophic forgetting \cite{kirkpatrick2017overcoming,lee2017overcoming}. We achieve some improvement when $L_D$ is added as an IPP in LwF-MC. The performance further improves with the addition of $L_{AD}$ in LwM configuration.

\section{Conclusion and future work}
We explored the IL problem for the task of object classification, and proposed a technique: LwM by combining $L_D$ with $L_{AD}$, for utilizing attention maps to transfer the knowledge of base classes from the teacher to student model, without requiring any data of base classes during training. This technique outperforms the baseline in all the scenarios that we investigate. Regarding future applications, LwM can be used in many real world scenarios. While we explore IL problem for classification in this work, we believe that the proposed approach can also be extended for segmentation. Incremental segmentation is a challenging problem due to the absence of abundant ground truth maps. The importance of incremental segmentation has already been underscored in \cite{baweja2018towardsNIPSW}. As visual attention is also meaningful for segmentation (as shown in \cite{Li_2018_CVPR}), we intend to extend LwM to incremental segmentation in the near future.

\section*{Acknowledgment}
Mostly done when Prithviraj was an intern at Siemens, this work was partially supported by the Intelligence Advanced Research Projects Activity (IARPA) via Department of Interior/Interior Business Center (DOI/IBC) contract number D17PC00345. The U.S. Government is authorized to reproduce and distribute reprints for Governmental purposes not withstanding any copyright annotation thereon. The views and conclusions contained herein are those of the authors and should not be interpreted as necessarily representing the official policies or endorsements, either expressed or implied of IARPA, DOI/IBC or the U.S. Government.

\end{document}